\documentclass[journal]{IEEEtran}
\ifCLASSINFOpdf
\else
\fi
\usepackage{multicol,lipsum}
\usepackage{cuted}
\usepackage{flushend}
\usepackage{amsmath, commath}
\usepackage{epsfig, graphicx, subfigure}
\usepackage{multirow}
\usepackage{hyperref}
\usepackage{makecell}
\usepackage{amssymb}
\hypersetup{
    colorlinks=true,
    linkcolor=blue,
    filecolor=magenta,      
    urlcolor=cyan,
}

\begin{document}
%
\title{An End-to-End Neural Network for Image Cropping by Learning Composition from Aesthetic Photos}

%
%
%

\author{Peng~Lu,
        Hao Zhang,
        Xujun~Peng,
        and Xiaofu~Jin
\thanks{P. Lu, H. Zhang and X. Jin are with the School of Computer Science, Beijing University of Posts and Telecommunications, Beijing, China. E-mail: plu@bupt.edu.cn}
\thanks{X. Peng is with the Information Sciences Institute, University of Southern California, Marina del Rey, CA, USA. E-mail: xpeng@isi.edu}
}

\maketitle

\begin{abstract}
As one of the fundamental techniques for image editing, image cropping discards unrelevant contents and remains the pleasing portions of the image to enhance the overall composition and achieve better visual/aesthetic perception. In previous work, we have shown that the regression network can help to obtain high aesthetic quality cropping images with multiple individual steps, and we are now expanding this work to an end-to-end approach in a probabilistic framework. In this paper, we primarily focus on improving the accuracy of automatic image cropping, and on further exploring its potential in public datasets with high efficiency. From this respect, we propose a deep learning based framework to learn the objects composition from photos with high aesthetic qualities, where an anchor region is detected through a convolutional neural network (CNN) with the Gaussian kernel to maintain the interested objects' integrity. This initial detected anchor area is then fed into a light weighted regression network to obtain the final cropping result. Unlike the conventional methods that multiple candidates are proposed and evaluated iteratively, only a single anchor region is produced in our model, which is  mapped to the final output directly. Thus, low computational resources are required for the proposed approach. The experimental results on the public datasets show that both cropping accuracy and efficiency achieve the state-of-the-art performances.
\end{abstract}

\begin{IEEEkeywords}
Deep learning, aesthetics, image composition, convolutional network.
\end{IEEEkeywords}

%
\IEEEpeerreviewmaketitle

\section{Introduction}
%
%
%
%
Image cropping, which aims at removing unexpected regions and non-informative noises from a photo/image, by modifying its aspect ratio or through improving the composition, is one of the basic image manipulation processes for graphic design, photography and image editing. Nowadays, with the  proliferation of hand-held smart devices, users are more eager to capture photos obtaining not only the theme that the image needs to express but also the good composition and appealing colors, to facilitate semantic searching and to make audiences enjoyable. This trend attracts increasing interests of image cropping from both research community and industries. \par

However, cropping an image to obtain appropriate composition for achieving better visual quality is notoriously difficult, primarily driven by three facts: (1) to determine the main object/theme of a given image is a nontrivial task, which needs deep domain knowledges and sophisticate skills; (2) assessment of aesthetic of the cropped image is highly subjective such that different viewers might have various opinions for the same cropped photo, or even the same viewer might have opposite feelings for the same image at different time; (3) vast amount of cropping candidate areas can be extracted from the image which causes the solution space is exponentially increased. \par

To tackle this problem, many researchers seek to propose novel approaches to automatically crop images with high aesthetic score. These existing researches can be roughly grouped into two main categories:  \textit{sliding-judging} based methods and \textit{determining-adjusting} based methods \cite{8365844}. \par

The sliding-judging based approaches normally exhaustively scan the entire image using windows with different size and aspect ratio to produce abundant candidate regions \cite{Fang:2014:AIC:2647868.2654979, Nishiyama:2009:SPC:1631272.1631384}. For each candidate, a classifier or ranker is applied to evaluate its visual/aesthetic quality and the one with the highest score is considered as the optimal cropping result. However, the low computation efficiency of these approaches limited their success. In order to avoid greedily searching against all possible sub-windows,  determining-adjusting based approaches attempt to propose a small set of candidate windows with high probabilities to narrow down the searching space for the optimal cropping rectangle. Normally, a seed candidate is initially determined by face, salient object, or attention detection algorithms \cite{Suh:2003:ATC:964696.964707, 7780430, 8237502}, from which the surrounding areas are scanned and evaluated by classifiers or rankers to select the region with the highest visual/aesthetic quality. Although determining-adjusting based approaches have higher efficiency than sliding-judging based methods, they still encounter the same problems of multiple candidates generation and selection. \par


\begin{figure}[hbt]
	\centering
    \subfigure[The proposed pipeline for image cropping.]{\includegraphics[width=0.8\columnwidth,clip]{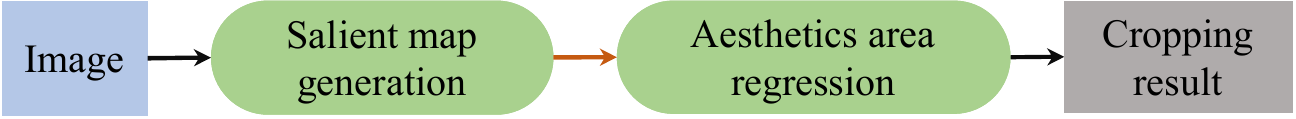} \label{fig:pipeline1}} 	
    \subfigure[Examples of the pipeline's intermediate outputs.]{\includegraphics[width=0.95\columnwidth,clip]{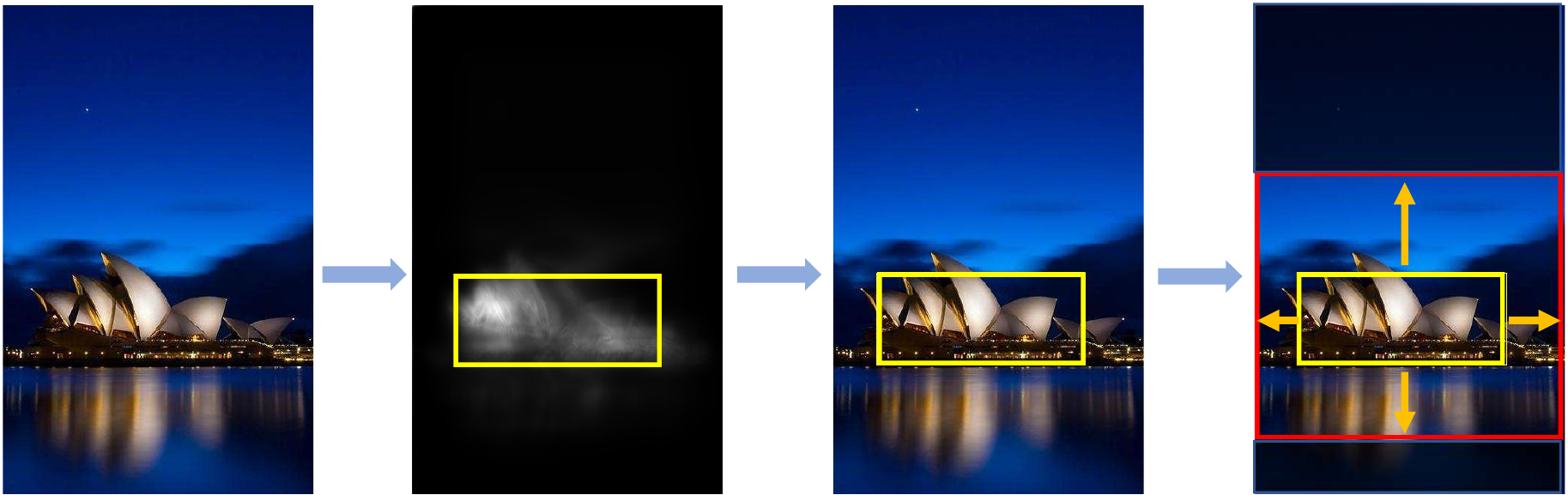} \label{fig:pipeline2}}
    \caption{The flowchart of the proposed image cropping system. (a) The overall pipeline of the proposed image cropping approach. (b) Examples of intermediate outputs of the proposed image cropping pipeline. }
\end{figure}

Thus, in this paper, we propose a new image cropping framework which finds the hidden relation between the interested objects and the areas with high aesthetics scores and is applied to determine the final cropping area directly. Particularly, the overall pipeline of the proposed image cropping system is illustrated in Figure \ref{fig:pipeline1}, where a deep neural network is applied to extract the saliency map of the image, which is followed by the proposed object integrity constraint layer to determine the anchor region that contains the interested objects in the image. Then a regression network is employed to reveal the relationship between the interested objects and the area with high aesthetic qualities and obtain the final cropping rectangle. The example of intermediate outputs for the proposed image cropping pipeline are demonstrated in Figure \ref{fig:pipeline2}. As can be observed from these figures, the proposed cropping method only has one pass to achieve the optimal cropping result, without iterative searching or scanning on multiple proposals of different windows, which highly improves the computational efficiency and obtains comparative accuracy performance.

In summary, we make three contributions to the literature:
\begin{itemize}
	\item Unlike our previous work that used multiple individual steps to accomplish the image cropping tasks \cite{LU20191}, we proposed a probabilistic framework to crop the image where an end-to-end deep neural network was implemented to facilitate the training and inference process, which outperformed the state-of-the-art methods;
    \item We revealed the relevance between salient objects and image's composition through convolutional neural network (CNN) and shedded light into the black box of neural networks to unveil the underlying principles between image composition and aesthetic, which avoided the candidates searching and evaluating steps to boost the system's efficiency and obtained the processing speed of 50 fps;
    \item A Gaussian-like kernel layer was proposed in this work to identify an anchor region from saliency map to include the interested objects in the image, which can be easily integrated into the neural network to help the searching of the optimal cropping area with high aesthetic scores. Compared to the salient rectangle proposal method employed in \cite{LU20191}, which was a stand-alone component introduced from \cite{7780430}, the proposed Gaussian-like kernel was more efficient for both training and inference.
\end{itemize}

The remainder of this paper is organized as follows. Section \ref{sec:previous_work} briefly covers the related works to image cropping. Section \ref{sec:proposed} is an in-depth introduction of the proposed methodology. The experimental setup, results analysis and discussion are presented in section \ref{sec:experiments}. Finally, we conclude our work in section \ref{sec:conclusion}.

\section{Previous Work} \label{sec:previous_work}
\subsection{Salient Object Detection}
Most vertebrates have the ability to move their eyes and predict fixation with limited time and resources, which enables them to focus on the most informative region and extract the most relevant features for the particular scene. This phenomenon inspired researchers to obtain the cropping areas of image relied on the salient object prediction. \par

Generally, saliency map is produced prior to the salient object detection, as demonstrated in \cite{6909437}, where each pixel in the map indicates the confidence of the fixation. In \cite{Harel:2006:GVS:2976456.2976525}, Harel \emph{et al.} proposed a graph-based visual saliency model depended on Markovian chain assumption, which calculated and normalized the activation map by measuring the dissimilarity between neighboring pixels. The reported ROC curve for this method beat the classical attention map detection approach proposed by Itti \emph{et al.}, where multiple empirical features were fed into a neural network to select the proper attended locations \cite{730558}. In the same manner, Judd \emph{et al.} defined a set of hand-crafted features to represent low-, mid- and high-level perception of human visual system and fed them into a support vector machine (SVM) to distinguish positive and negative salient pixels \cite{5459462}. 

However, the drawback of these mentioned approaches is that strong domain knowledge and experiences are required for design of those hand-tuned features, which is a obstacle to extend their applicability. Therefore, in \cite{6909754}, Vig \emph{et al.} proposed to utilize CNN to learn the representations for salient and non-salient regions. With labeled feature vectors, an L2-regularized, linear, L2-loss SVM was trained to predict the probability of fixation of images in their work. Similarly, Pan \emph{et al.} introduced two CNN based neural networks to accomplish the saliency map generation task, where the shallow CNN's outputs were mapped from its last fully connected layer, and the deep CNN's outputs were generated from a deconvolution layer \cite{7780440}. \par

By arguing that the problem definition of conventional salient object extraction approaches is ill-posed because various observers might have different views for what constitutes a salient object, Islam \emph{et al.} re-modeled salient object detection as a relative rank problem \cite{8578844}. In their framework, a coarse nested relative salience stack was generated through a CNN based encoder, which was successively refined by a stage-wise refinement network where rank-aware refinement units were applied.

\subsection{aesthetic Assessment}
Besides salient objects that affect the performance of image cropping, aesthetic, which represents  the  degree  of  beauty, is another key factor to determine the quality of cropped images. Early work for aesthetic assessment can be dated back to the researches of color harmony theories \cite{Moon:44} and photographic composition \cite{grill:composition}. In recent years, many automatic image aesthetic assessment algorithms were proposed, where hand-crafted global features, such as spatial distribution of edges, color distributions, hue count etc. \cite{1640788} and local features, e.g. wavelet-based texture and  shape convexity \cite{10.1007/11744078_23} were employed. To take the advantage of both global and local features, Zhang \emph{et al.} combined structural cues of these two levels for photo aesthetic evaluation \cite{6728663}. Particularly, graphlet-based local structure descriptors were constructed and projected onto a manifold to preserve the global layout of the image, which was embedded into a probabilistic framework to assess image aesthetic. However, these representations consider  whole  image indiscriminately ignoring  the importance of main subjects in the image. To remedy this problem, Luo \emph{et al.} suggested extracting different subject areas prior to the aesthetic evaluation and treating them using different aesthetic features \cite{6544270}. Furthermore, genetic image descriptors were also applied to facilitate the aesthetic assessment task. For instance, Marchesotti \emph{et al.} developed two types of local image signatures originated from Bag-of-visual-words and Fisher vectors by incorporating SIFT and color information into them \cite{6126444}. \par

Under the assumption that semantic recognition task can help the aesthetic assessment, Kao \emph{et al.} proposed a multi-task framework where two tasks were trained simultaneously while the  representations were shared by two networks \cite{7814292}. This idea was also applied by Lu \emph{et al.} in their work of color harmony modeling, which used both bag-of-visual-words features and semantic tag information to boost the aesthetic assessment performance through colors \cite{LU2016731}. Moreover, in order to overcome the problem of contaminated tags, a semi-supervised deep active learning algorithm was proposed in \cite{8340874}, where a large set of object patches were extracted and linked to the semantical tags to benefit image aesthetic assessment. Regarding the professional photographer tend to employ different rules for image capturing according to the visual content, Tian \emph{et al.} designed a query-dependent aesthetic assessment method where each image was associated with a unique aesthetic model instead of the universe model based on its visual/semantics properties \cite{7271097}. 

\subsection{Regression Networks}
Although photos can be cropped depended on the obtained salient objects only, they are not necessary to be with high aesthetic quality  because the aspect of aesthetic is normally ignored for saliency detection. To tackle this problem, one feasible solution is to determine a seed cropping window according to the saliency map and propose a set of candidates around this seed window. Thereafter, every candidate is evaluated by its aesthetic quality and the one with the highest aesthetic score is considered as the final cropping result, as the methods described in \cite{Nishiyama:2009:SPC:1631272.1631384, 6618974}. However, iterative assessment for each candidate window's aesthetic score increases the computational complexity. Thus, a more practical and efficient approach is to map the seed salient region to the final cropping window directly using regression models, where the aesthetic information is integrated into the system. \par

In the field of computer vision, regression method is widely used for object detection. Girshick \emph{et al.} combined regions with CNN features (R-CNN) to find different objects in the image \cite{6909475}, where bounding-box regression technique proposed in \cite{5255236} was applied for a selective search region proposal to refine the detection window . To speed up R-CNN, Girshick improved their work by using a fully connected network to predict the bounding-box regression offsets and confidence of each proposal \cite{7410526}. Instead of performing classification for detection problem, Redmon \emph{et al.} framed object detection as a regression problem, where the input image was divided into small patches initially and the bounding boxes offsets as well as their probabilities for each class were predicted directly in one neural network, which was called YOLO \cite{7780460}. The final detections were obtained by merging bounding boxes for the same class. To make YOLO better and faster, Redmon and Farhadi shrink the CNN and used region proposal network (RPN) to generate more anchor boxes for boosting recall and localization accuracy, where regression network was remained for location/confidence prediction \cite{8100173}. Unlike YOLO, Liu \emph{et al.} detected different objects by evaluating a small set of boxes that were produced through multi-resolution CNN feature maps. The final bounding boxes of objects were also obtained by regressing to offsets for the centers of the default boxes \cite{10.1007/978-3-319-46448-0_2}. The similar ideas of using regression networks for objects detection can be found in \cite{8237586, 8099589}.

\subsection{Image Cropping \& Recomposition}
As an important procedure to enhance the visual quality of photos, image cropping and recomposition benefit from the development of salient object detection, aesthetic assessment and other computer vision techniques. To combine visual composition, boundary simplicity and content preservation into a photo cropping system, saliency map and salient object were used to encode the spatial configuration and content information, and gradient values were applied to measure the simplicity of image  by Fang \emph{et al.} \cite{Fang:2014:AIC:2647868.2654979}. In this method, image was densely cropped, evaluated and merged by the mentioned features to obtain the optimal cropping results. By segmenting the entire image into small regions, a region adjacency graph (graphlets) was constructed by Zhang \emph{et al.} to represent the aesthetic features of the image, from which the image was cropped through a  probabilistic model \cite{6327366, 6644258}. Zhang \emph{et al.} also extended the idea of graphlets in the semantic space for image cropping, which was created based on the category information of the images \cite{6766723}. In the semantic space, semantically representative graphlets were selected sequentially and evaluated by a pre-trained aesthetic prior model to guide the cropping process. Unlike the other algorithms that evaluated multiple candidate cropping areas, Samii \emph{et al.} searched against a high quality image database to find exemplar photos with similar spatial layouts as the query image, and matched the composition of the query image to each of exemplars by minimizing composition distance in a high-level context feature space to calculate the optimal crop areas \cite{Samii:2015:DAC:2771690.2771704}. In \cite{7045920}, Wang \emph{et al.} applied similar  concept  for photo cropping that exploited sparse auto-encoder to discover the composition basis from a database containing well-composed images. Differ from \cite{Samii:2015:DAC:2771690.2771704}, Wang's method organized the learning and inference in a cascade manner to achieve higher efficiency. By considering that perspective effect is one of the most commonly used techniques for photography, Zhou \emph{et al.} developed a hierarchical segmentation method integrating photometric cues with perspective geometric cue to detect the dominant vanishing point in the image, which was employed for image re-composition or cropping \cite{Zhou:2015:MPE:2733373.2806248}.

  \begin{figure*}[htp]
      \centering
      \includegraphics[width=0.8\textwidth,clip]{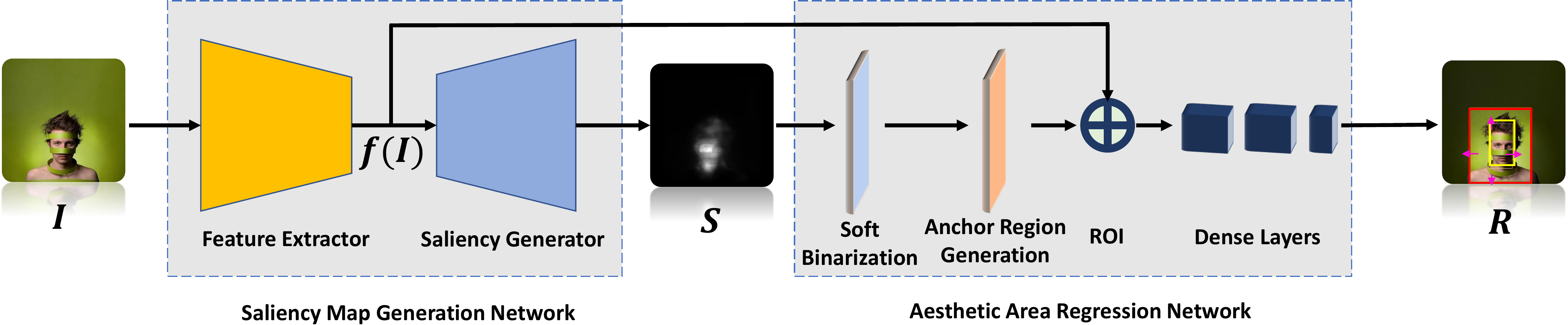}  
      \caption{Architecture of the proposed saliency map detection and aesthetic area regression network. \label{fig:structure}}
  \end{figure*}

Recently, thanks to the development of DNN, the research of image cropping tends to utilize deep learning approaches. By imitating the process of professional photographic, Chen \emph{et al.} proposed a ranking CNN to harvest unambiguous pairwise aesthetic ranking examples on the web and applied this network to find the optimal cropping result from many candidate regions \cite{Chen:2017:LCP:3123266.3123274}. Instead of generating the attention map for cropping, Kao \emph{et al.}   proposed to use aesthetic map, which was extracted via a CNN, and gradient energy map to accomplish the image cropping task, by learning the composition rules through a SVM classifier \cite{7952503}. In \cite{8259308}, Guo \emph{et al.} designed a cascaded cropping regression (CCR) approach to crop the image, where a deep CNN was applied to extract features from images and the cropping areas were predicted by the proposed CCR algorithm. Inspired by human's decision making, Li \emph{et al.} designed a weakly supervised aesthetic aware reinforcement learning framework to address the problem of image cropping, where the photo was initially cropped and repeatedly updated based on the current observation and the historical experience \cite{8578953}. In \cite{LU20191}, Lu \emph{et al.} proposed a regression network based cropping method, which mapped initial detected saliency rectangle to a cropping area with high aesthetics quality. Unlike the conventional photo cropping method that only produced a single output, in \cite{wei2018good}, Wei \emph{et al.} proposed a system that returned multiple cropping outputs based on a teacher-student framework. In this framework, the teacher was trained to evaluate candidate  anchor boxes, and the scores from the teacher were used to supervise the training of student, a view proposal net. The combination of these two networks effectively improve the cropping performance. The interest readers can refer \cite{Islam:2017:SAI:3077426.3077444} for more comprehensive surveys.

\section{Proposed Approach} \label{sec:proposed}
\subsection{Motivation \& System Overview} \label{sec:motivation}

By studying the procedure of professional photography, we can see that the theme is firstly determined by the photographer prior to other actions. To express this theme, the objects along with the compatible backgrounds are selected subsequently. Once the main objects contained in  the  photo  are  given, the other parameters for the photography, such as exposure time,  composition,  colors,  etc., will be set for the final shooting. \par

    \begin{figure*} 
        \centering
        \includegraphics[width=0.8\textwidth,clip]{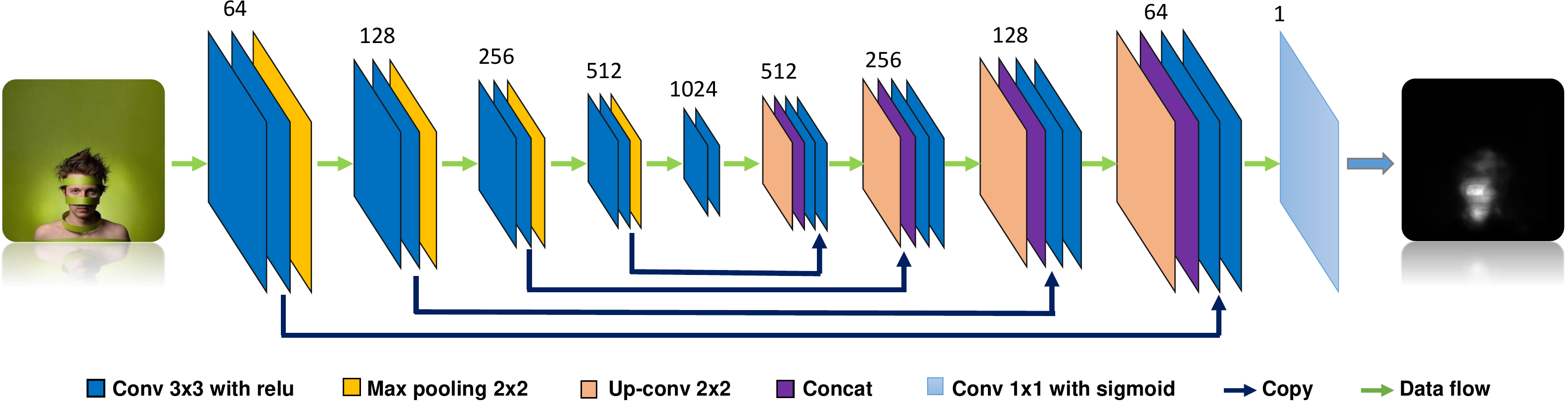} 
        \caption{The U-Shaped network implemented in this work for saliency map detection. \label{fig:unet}}
    \end{figure*}

Based on this observation, the process of image cropping to obtain the high aesthetic quality can be decomposed into two steps: detection of the interested objects $\mathcal{S}$ in the image $\mathcal{I}$ and prediction of the aesthetic areas $\mathcal{R}$ of the image based on the objects of interest $\mathcal{S}$. This process can be formally expressed as:
    \begin{equation}
        P(\mathcal{R, S | I}) = P(\mathcal{S | I}) \cdot P(\mathcal{R | S, I}),
    \end{equation}
where $ \mathcal{S}=\{S_{i,j}| i \times j \in |\mathcal{I}| \}$ denotes the interested objects of the image, $|\mathcal{I}|$ means the number of pixels in the image, and $\mathcal{S}_{i,j} \in \{0,1\}$ represents whether a given pixel belongs to the objects of interest. $P(\mathcal{S} | \mathcal{I})$ is the probability of interested objects $\mathcal{S}$ given an image $\mathcal{I}$, and $P(\mathcal{R}|\mathcal{S,I})$, which reveals the hidden relationship between the interested objects and the final cropping region, denotes the probability of $\mathcal{R}$ with respect to the image $\mathcal{I}$ and the detected objects of interest $\mathcal{S}$.

Thus, the aesthetic region of an image can be obtained if $P(\mathcal{R,S|I})$ is calculated. Hence, a probabilistic model based cropping system can be designed whose parameters can be expressed as $\mathbf\Theta$, and the image cropping task can be considered as the maximum likehood (ML) estimation problem
for a given training image $\mathcal{I}_i$, along with its ground truths $\mathcal{S}_i$ and $\mathcal{R}_i$:

    \begin{align} \label{eq:system}
        \nonumber \mathbf\Theta & = \arg\max_{\mathbf\Theta} \sum_{i=1}^N P(\mathcal{R}^{(i)},\mathcal{S}^{(i)}|\mathcal{I}^{(i)}; \mathbf\Theta) \\
        \nonumber & = \arg\max_{\mathbf\Theta} \sum_{k=1}^N P(\mathcal{S}^{(k)}|\mathcal{I}^{(k)}; \mathbf\Theta_s) \cdot P( \mathcal{R}^{(k)}|\mathcal{S}^{(k)}, \mathcal{I}^{(k)}; \mathbf\Theta_r) \\
        \nonumber &  = \arg\max_{\mathbf\Theta} \sum_{k=1}^N \big ( \log P(\mathcal{S}^{(k)}|\mathcal{I}^{(k)}; \mathbf\Theta_s) \\
            & \phantom{{}=1} + \log P(\mathcal{R}^{(k)}|\mathcal{S}^{(k)}, \mathcal{I}^{(k)}; \mathbf\Theta_r) \big ) ,
    \end{align}
where superscript $k$ indicates the index of training sample and ground truth, $N$ is the total number of training samples, and $\mathbf\Theta=[\mathbf\Theta_s, \mathbf\Theta_r]^T$ are the parameters of the model. \par



Based on this analysis, we design an end-to-end DNN based image cropping system that follows the probability framework as described in Eq. \ref{eq:system}. In the proposed cropping system, two main components are constructed, where the saliency map  generation network $H(\mathcal{I};\mathbf\Theta_s)$ in the Fig. \ref{fig:structure} is served to predict $\mathcal{S}$ given image $\mathcal{I}$. And aesthetic area regression network $G(\mathcal{I},\mathcal{S};\mathbf\Theta_r)$ containing the proposed anchor region generation layer, ROI pooling layer and fully connected layers is used as a regressor to produce final cropping outputs $\mathcal{R}$ based on $\mathcal{I}$ and $\mathcal{S}$.  These two components are corresponding to the photographer's actions of the objects decision and final cropping areas selection. \par

Thus, maximizing the Eq. \ref{eq:system} is equivalent to minimizing the loss $L_{total}$ of the neural network:
    \begin{align}
        \nonumber \mathbf\Theta & = \arg\min_{\mathbf\Theta} \mathcal{L}_{total} \\
        & = \frac{1}{N} \sum_{k=1}^N \arg\min_{\mathbf\Theta} \left ( \mathcal{L}_s (\mathcal{\hat{S}}^{(k)}, \mathcal{S}^{(k)} )  + \lambda \mathcal{L}_r(\mathcal{\hat{R}}^{(k)}, \mathcal{R}^{(k)}) \right ),
    \end{align}
where $\mathcal{L}_s(\cdot)$ represents the loss from the inconsistency between predicted $\mathcal{\hat{S}}^{(k)}$ by the saliency map detection network and ground truth $\mathcal{S}^{(k)}$ of the images $\mathcal{I}^{(k)}$, $\mathcal{L}_r(\cdot)$ is the loss caused by the difference between predicted aethetic region $\mathcal{\hat{R}}^{(k)}$, and the ground truth region $\mathcal{R}^{(k)}$, and $\lambda$ is the weight controlling the influence from these two networks and we use $\lambda=1$ in this work.    


As can be seen from the Fig. \ref{fig:structure}, unlike the conventional image cropping methods that explicitly or implicitly generate and evaluate multiple candidate cropping regions, the proposed system takes the input image to extract the salient area and maps this single area to the final output region by regression network directly. Thus, in this framework the data flows through the network only once without extensively assessing multiple candidates, which highly improves the efficiency and maintains the accuracy in the meantime. \par


\subsection{Saliency Map Generation Network}

Saliency map detection aims at predicting visually salient objects in an image that attract human attention. To mimic the professional photography step of choosing the main objects that express the theme of photos, it is feasible to generate a saliency map for image cropping which contains the interested objects in the image.


In the proposed system, we adopt a modified U-shaped network to produce the salicency map. As a variant of widely used fully convolutonal encoder-decoder, U-shaped network is originally designed for semantic segmentation on biomedical images \cite{10.1007/978-3-319-24574-4_28}. It merges feature maps from convolutional layers to deconvolutional layers gradually during the upsampling procedure. Thus, different types of features are preserved for the semantic labeling task. \par

Particularly, in our implementation, the encoder for the U-shaped network is composed by four fundamental blocks, where every two convolutional layers followed by a max pooling layer are stacked to form the basic block. Similarly, a decoder is constructed by four basic blocks where every two deconvolutional layers and a upsampling layer are used. For each fundamental block in the encoder, its feature maps are copied and concatenated directly to the corresponding block in the decoder with the same size of feature dimensions. Thus, the encoder is employed to extract features for the image and the saliency map is produced based on the decoder.  The detailed structure of the U-shaped network we implemented is illustrated in Fig. \ref{fig:unet}. \par


\subsection{Aesthetic Area Regression Network}
Based on the detected saliency map, the relation between the interested objects and the area with high aesthetic quality can be learned through the proposed anchor region detection layer and regression layers, which are described in the following subsections with details.

\subsubsection{Soft Binarization Layer} \label{sec:enhancement}

To make our cropping system less sensitive to the presence of outliers in the saliency map, we introduce a  function $\rho(x; \sigma)$ to enhance the quality of interested objects in saliency map, which is defined by:
\begin{equation}
    \rho(x; \sigma) = \frac{x^2}{x^2+\sigma^2} .
\end{equation}

\begin{figure}[h]
    \centering
    \subfigure[]{\includegraphics[width=0.3\columnwidth, clip]{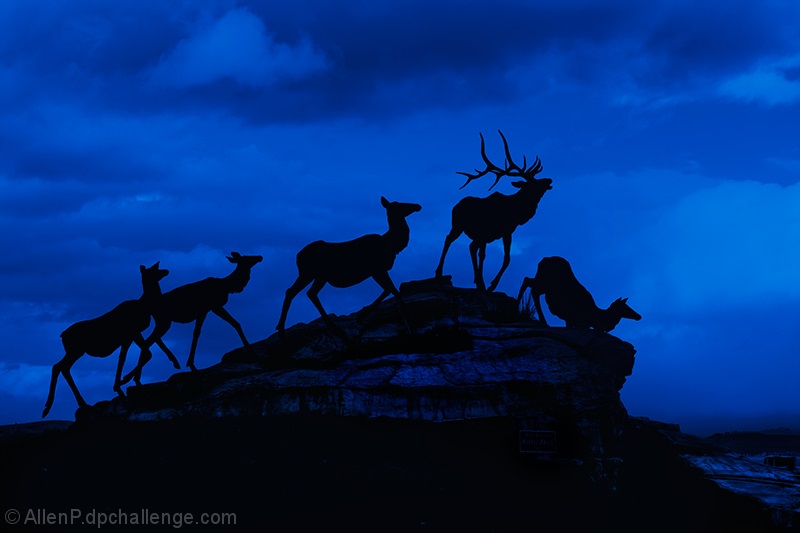}}
    \subfigure[]{\includegraphics[width=0.3\columnwidth, clip]{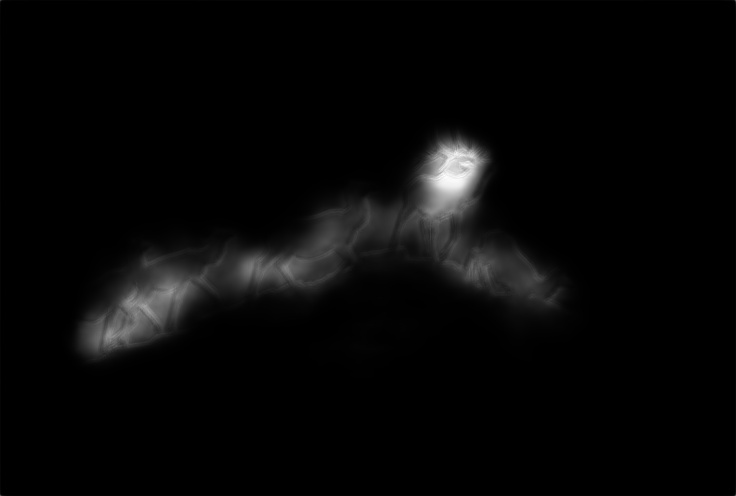}}
    \subfigure[]{\includegraphics[width=0.3\columnwidth, clip]{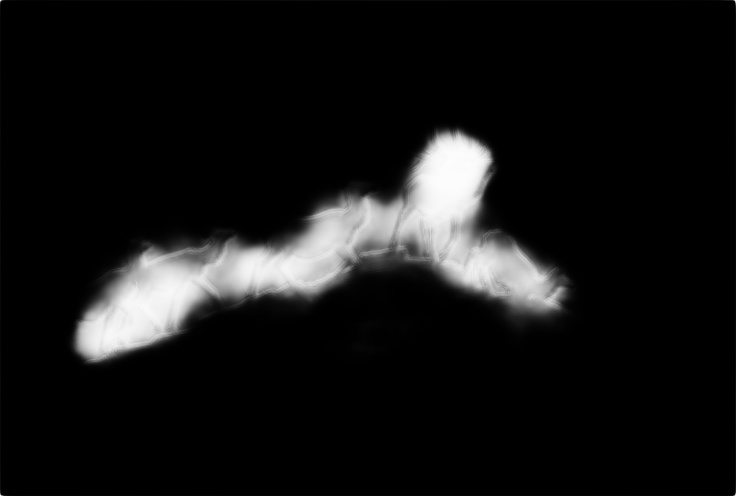}}
    \caption{Sample saliency map and enhanced salient object image for color image with high aesthetic scores. (a) The original high quality image. (b) The corresponding saliency map. (c) Enhanced saliency map by using function $\rho(x; \sigma)$, where the interested objects are easily distinguished from backgrounds. \label{fig:enhanced-saliency}}
\end{figure}

By selecting proper scale parameter $\sigma$, $\rho(x; \sigma)$ function maps small value of pixel in saliency map to 0 and will be saturated to 1 with larger pixel value, given a normalized saliency map whose values are between 0 and 1. In Fig. \ref{fig:enhanced-saliency}, we demonstrate a sample saliency map and its corresponding enhanced version, from which we can observe that the difference between the interested objects in the image and the background is enlarged and they can be distinguished with minimum efforts. \par

As the derivative of $\rho(x; \sigma)$ function is calculated as:
\begin{equation}
    \frac{\partial\rho(x; \sigma)}{\partial x} = \frac{2x\sigma^2}{(x^2+\sigma^2)^2} ,
\end{equation}
this operation can be easily integrated into the proposed neural network pipeline without blocking the backpropagation.

\subsubsection{Anchor Region Generation Layer} 
Based on the obtained soft binarization saliency map $\mathcal{S}$ that shows each pixel's probability to be the interested objects, it is necessary to model the $P(\mathcal{R} | \mathcal{S, I})$ to reveal the relations between the interested objects and the final cropping window. In order to represent this relation, it needs to extract the features of the interested objects first. A feasible approach is to find an anchor region from $\mathcal{S}$ which is not only deterministically determined by the interested objects in the saliency map but exclusively encloses them, and use the features from $\mathcal{I}$ within this anchor region to represent interested objects features. Then, the final cropping area with high aesthetic scores $\mathcal{R}$ can be inferred from features in anchor region. 

Inspired by the mean shift algorithm that is used to locate and track the face regions in videos \cite{Bradski98computervision}, a region generation algorithm is proposed in this work to perform the initial anchor salient region creation. \par

Given a saliency map $\mathcal{S}$ extracted by U-shaped network, the center of mass $(c_x, c_y)$ for this map  can be calculated by:
	\begin{equation}
	\nonumber	c_x = \frac{M_{10}}{M_{00}}, \hspace{3mm}  c_y = \frac{M_{01}}{M_{00}},
	\end{equation}
and the standard deviation for the center of mass are obtained according to:
	\begin{align}
    \nonumber \sigma_x = \sqrt{\frac{M_{20}}{M_{00}} - c^2_x}, \hspace{3mm} 
    \sigma_y = \sqrt{\frac{M_{02}}{M_{00}} - c^2_y},
    \end{align}
where moments $M_{00}$, $M_{01}$, $M_{10}$, $M_{20}$ and $M_{02}$ are calculated based on:
	\begin{align}    	
		M_{00} & = \sum_{i,j}S_{i,j} \\
        M_{10} & = \sum_{i,j} i \cdot S_{i,j}, \hspace{3mm} M_{01} = \sum_{i,j} j \cdot S_{i,j} \\
        M_{20} & = \sum_{i,j} i^2 \cdot S_{i,j}, \hspace{3mm} M_{02} = \sum_{i,j} j^2 \cdot S_{i,j}.
	\end{align}

Therefore, a region that includes the energy of the saliency map can be defined through its top right corner $(x^s_{tr}, y^s_{tr})$ and bottom left corner $(x^s_{bl}, y^s_{bl})$ by using a Gaussian-like window:
  \begin{equation}
      (x^s_{tr}, y^s_{tr})  = (c_x + \gamma \sigma_x, c_y + \gamma \sigma_y) \label{eq:tr}
  \end{equation}
and
  \begin{equation}
      (x^s_{bl}, y^s_{bl})  = (c_x - \gamma \sigma_x, c_y - \gamma \sigma_y), \label{eq:bl}
  \end{equation}
where $\gamma$ is a hyper-parameter controlling the amount of energy contained in the window and maintaining the integrity of interested objects in the image. In this work, $\gamma=3.0$ is applied to include $99\%$ energy from the interested objects in the image. \par

To allow backpropagation of the loss pass through this region generation layer, the gradient of the cropping window's coordinates with respect to $\mathcal{S}$ can be defined. For coordinate $x$ of top right corner,  the partial derivative is given by:
	\begin{equation} \label{eq:derivative}
    	\frac{\partial x^s_{tr}}{\partial S_{i,j}} = \frac{\partial c_x}{\partial S_{i,j}} + \gamma \frac{\partial \sigma_x}{\partial S_{i,j}},
    \end{equation}
where $\frac{\partial c_x}{\partial S_{i,j}}$ and $\frac{\partial \sigma_x}{\partial S_{i,j}}$ are further calculated based on following equations:
    \begin{align} \label{eq:derivative1}
    	\nonumber \frac{\partial c_x}{\partial S_{i,j}} & = \frac{1}{M_{00}} \frac{\partial M_{10}}{\partial S_{i,j}} - \frac{M_{10}}{M^2_{00}} \frac{\partial M_{00}}{\partial S_{i,j}} \\
        & = \frac{i}{M_{00}} - \frac{M_{10}}{M^2_{00}}
    \end{align}
and
	\begin{align} \label{eq:derivative2}
    	\nonumber \frac{\partial \sigma_x}{\partial S_{i,j}} & = \frac{1}{2\sqrt{\frac{M_{20}}{M_{00}} - c^2_x}} \\
        \nonumber & \hspace{5mm} \times \bigg ( \frac{1}{M_{00}} \frac{\partial M_{20}}{\partial S_{i,j}} - \frac{M_{20}}{M^2_{00}} \frac{\partial M_{00}}{\partial S_{i,j}} - 2 c_x \frac{\partial c_x}{\partial S_{i,j}}  \bigg ) \\
        \nonumber & = \frac{1}{2\sqrt{\frac{M_{20}}{M_{00}} - c^2_x}} \\
        & \hspace{5mm} \times \bigg \{ \frac{i^2}{M_{00}} - \frac{M_{20}}{M^2_{00}} - 2 c_x \bigg ( \frac{i}{M_{00}} - \frac{M_{10}}{M^2_{00}}   \bigg ) \bigg \}
    \end{align}
and similar partial derivatives can be applied for $\frac{\partial y^s_{tr}}{\partial S_{i,j}}$, $\frac{\partial x^s_{bl}}{\partial S_{i,j}}$ and $\frac{\partial y^s_{bl}}{\partial S_{i,j}}$.

This provides the proposed anchor region generation layer a mechanism that allows loss gradients to flow back to the input of network to update the model's parameters $\mathbf\Theta$. 

\subsubsection{Dense Layers}

Although the anchor region produced by the anchor region generation layer contains most significant objects in the image, it is mostly far from having high aesthetic quality. So, based on the observation that the professional photographer tends to adjust the scene area for final shooting according to the interested objects, and the discovery that  \textit{``one may still roughly infer the extent of an object if only the middle of the object is visible"} \cite{Ren:2017:FRT:3101720.3101780}, three fully connected layers that maps the anchor region to the eventual cropping window with high visual quality are implemented. \par

In our implementation, the region of interest (RoI) pooling layer followed by fully connected layers is used to estimate final cropping areas. The RoI pooling layer is proposed in \cite{NIPS2016_6465}, which takes two inputs: the coordinates of predicted anchor region and the feature maps generated from the bottle layer of U-shaped network. Prior to feeding into the RoI pooling layer, the coordinates of predicted anchor region are reduced 16 times to match the size of feature maps from bottle layer in U-shaped network. In the RoI pooling layer, only features from anchor region are extracted, which are consequently passed to two fully connected layers with ReLU activation, whose sizes are 2048 and 1024, respectively. The last layer of this regression network is a fully connected layer who has 4 units with linear activation function, which predicts the four coefficients defined by Eq. \ref{equ_coordinates_1} and Eq. \ref{equ_coordinates_2}. \par

\subsubsection{Aesthetic Area Representation} \label{sec:regression_net}

To represent the relation between the detected anchor region and areas with high aesthetic qualities, we used the approach described in \cite{LU20191}. Given a detected anchor region, whose size is $w^s \times h^s$, if its corresponding high aesthetic quality image's size is $w^a \times h^a$, and their top-left and bottom-right corners are $(x^s_{tl}, y^s_{tl})$, $(x^s_{br}, y^s_{br})$, $(x^a_{tl}, y^a_{tl})$ and $(x^a_{br}, y^a_{br})$, respectively,  the offsets between the corners of these two rectangles $R \big ((x^s_{tl}, y^s_{tl}), (x^s_{br}, y^s_{br}) \big )$ and $R \big ((x^a_{tl}, y^a_{tl}), (x^a_{br}, y^a_{br}) \big )$ can be represented as:
	\begin{align}
    	 (\Delta x_t, \Delta y_t) & = (x^s_{tl}, y^s_{tl}) - (x^a_{tl}, y^a_{tl}) \\
         (\Delta x_b, \Delta y_b) & = (x^a_{br}, y^a_{br}) - (x^s_{br}, y^s_{br}).
    \end{align}

Hence, the height and width of these two rectangles can be expressed as:
	\begin{equation}
      	h^a = h^s + \Delta y_t + \Delta y_b = h^s + \alpha_t \cdot h^a + \alpha_b \cdot h^a \label{equ_coordinates_1}           \end{equation}
and
    \begin{equation}
        w^a = w^s + \Delta x_t + \Delta x_b = w^s + \beta_t \cdot w^a + \beta_b \cdot w^a , \label{equ_coordinates_2} 
    \end{equation}
where $\mathcal{O}=[ \alpha_t, \alpha_b, \beta_t, \beta_b ]$ are four coefficients. \par

In our implementation, the above coefficients $\mathcal{O}=[ \alpha_t, \alpha_b, \beta_t, \beta_b ]$ are used to represent the final aesthetic area and can be learned through a neural network. \par

During the testing, the corner coordinates $(x^s_{tl}, y^s_{tl})$, $(x^s_{br}, y^s_{br})$ of the anchor region of input image and four coefficients $\mathcal{O}=[ \alpha_t, \alpha_b, \beta_t, \beta_b ]$ are predicted by the proposed end-to-end network. Then, the width and height of final aesthetic region can be expressed  as follows:
	\begin{align}
    	 h^a = \frac{ y^s_{br}-y^s_{tl} }{1 - \alpha_t -  \alpha_b} \\
    	 w^a = \frac{ x^s_{br}-x^s_{tl} }{1 - \beta_t - \beta_b} 
    \end{align}

Thus, the coordinates of top-left and bottom-right corners of aesthetic area can be calculated by:
	\begin{align}
    	\nonumber x^a_{tl} & = x^s_{tl} - \beta_t \cdot w^a \hspace{1cm}
    	y^a_{tl} = y^s_{tl} - \alpha_t \cdot h^a \\
    	\nonumber x^a_{br} & = x^s_{br} + \beta_b \cdot w^a  \hspace{1cm}
    	y^a_{br}  = x^s_{br} + \alpha_b \cdot h^a
    \end{align}

\subsection{Loss Functions for the Cropping System} \label{sec:loss}
As introduced in subsection \ref{sec:motivation}, the total loss of the proposed neural network based cropping system is given by:
    \begin{equation}
        \mathcal{L}_{total} = \frac{1}{N} \sum_{k=1}^N \left ( \mathcal{L}_s(\cdot) + \lambda \mathcal{L}_r(\cdot) \right )
    \end{equation}
where $\mathcal{L}_s(\cdot)$ is the loss from saliency map detection network and $\mathcal{L}_r(\cdot)$ is the loss from aesthetic regression network, $N$ means the total training number, and $\lambda$ is the weight controlling the influence from these two networks.



To train the U-shaped based network $H(\mathcal{I}, \mathbf\Theta_s)$, the binary cross-entropy of each pixel is calculated:
	\begin{align}
    	\nonumber  H \big (\mathcal{I}_{i,j}; \mathbf\Theta_s & =(\mathbf{W}_s,\mathbf{b}_s) \big ) \\
    	\nonumber & = - S_{i,j} \log p(\mathcal{I}_{i,j};(\mathbf{W}_s,\mathbf{b}_s)) \\
    	& - (1-S_{i,j}) \log \big ( 1-p(\mathcal{I}_{i,j};(\mathbf{W}_s,\mathbf{b}_s)) \big ),
    \end{align}
where $[\mathbf{W}_s, \mathbf{b}_s]$ are weights of U-shaped saliency map detection network, $p(\mathcal{I}_{i,j};(\mathbf{W}_s,\mathbf{b}_s))$ stands for the predicted confidence for the interested objects of each pixel, and $\mathcal{\hat{S}}_{i,j} = p \big (\mathcal{I}_{i,j};(\mathbf{W}_s,\mathbf{b}_s) \big )$ holds for the detected saliency map $\mathcal{\hat{S}}$.

Thereafter, the loss for a given image $\mathcal{I}^{(k)}$ can be expressed as:
	\begin{align}
    	\nonumber \mathcal{L}_s(\mathcal{\hat{S}}^{(k)},\mathcal{S}^{(k)} ) & = \mathcal{L}_s(\mathbf{W}_s,\mathbf{b}_s) \\
    	& = \sum_{\mathcal{I}^{(k)}_{i,j}} H \left (\mathcal{I}^{(k)}_{i,j}; (\mathbf{W}_s,\mathbf{b}_s) \right ),
	\end{align}
where superscript $k$ is the indexl of the training sample.

And as described in section \ref{sec:regression_net}, unlike other image cropping methods that train a ranker or classifier to evaluate the cropping areas' aesthetic quality by using training samples with high/low qualities, the proposed aesthetic region regression network uses a regressor to predict the cropping window, where only features from high aesthetic images are required and learned. Thus, in our training phase for the regression network, the anchor salient region $\mathcal{R} \big ((x^s_{tr}, y^s_{tr}), (x^s_{bl}, y^s_{bl}) \big )$ for high aesthetic quality image is firstly detected by anchor region generation layer. Then, the region of original high quality image $\mathcal{R} \big ((x^a_{tr}, y^a_{tr}), (x^a_{bl}, y^a_{bl}) \big )$ is used to calculate the offsets coefficients $\mathcal{O} = [ \alpha_t, \alpha_b, \beta_t, \beta_b ]$, where $(x^a_{tr}, y^a_{tr})=(0,0)$, $(x^a_{bl}, y^a_{bl}) = (w^a, h^a)$ and $w^a \times h^a$ is high quality image's size. And these offsets coefficients are used to supervise the training of the proposed regression network, where $L2$ loss is applied according to:
    \begin{align}
       \nonumber  \mathcal{L}_r(\mathcal{\hat{R}}^{(k)}, \mathcal{R}^{(k)}) & = \mathcal{L}_r(\mathbf{W}_r, \mathbf{b}_r) \\
        & = \big \| \mathcal{\hat{O}}^{(k)} - \mathcal{O}^{(k)} \big \|^2
    \end{align}
where $[\mathbf{W}_r, \mathbf{b}_r]$ are system weights for aesthetic regression network, $\mathcal{O}^{(k)}$ is the ground truth of offsets coefficients of an image $\mathcal{I}^{(k)}$ and $\mathcal{\hat{O}}^{(k)}$ is the corresponding predicted offsets coefficients. \par

The source code and the pretrained models of this work are available from: \url{https://github.com/CVBase-Bupt/EndtoEndCroppingSystem}.


\section{Experiments} \label{sec:experiments}
\subsection{Datasets and Evaluation Protocol} \label{sec:dataset}

\begin{figure*}
  \centering
    \subfigure[]{\includegraphics[width=0.16\textwidth,clip]{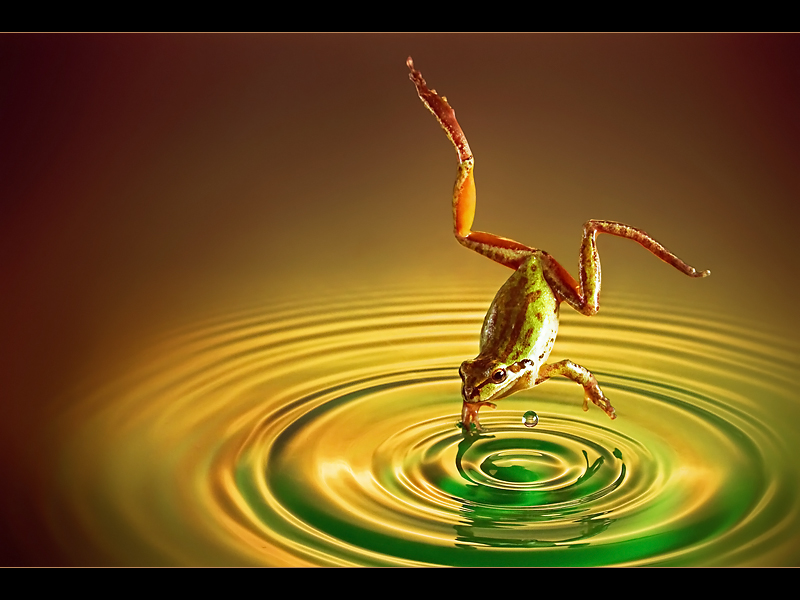} 
        \includegraphics[width=0.15\textwidth,clip]{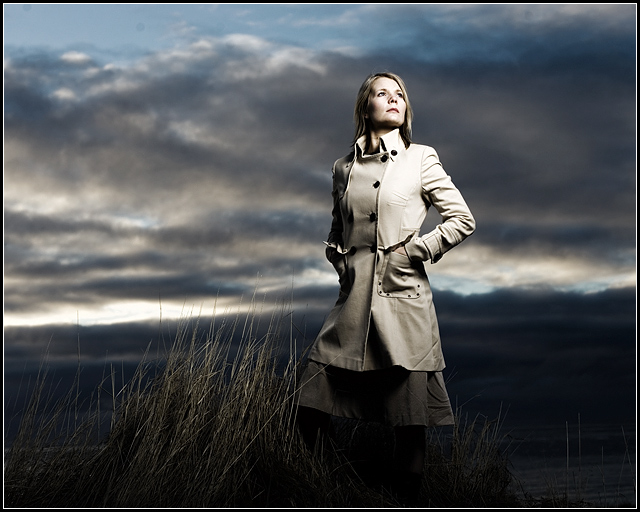} 
        \includegraphics[width=0.18\textwidth,clip]{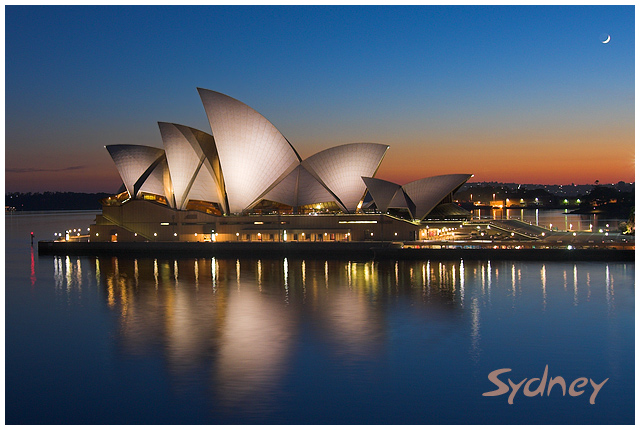}
        \includegraphics[width=0.15\textwidth, clip]{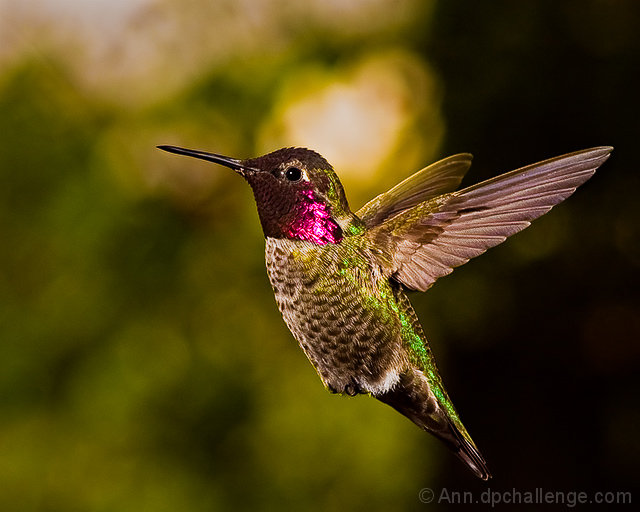} \label{fig:ava-origin}} \\	
    \subfigure[]{\includegraphics[width=0.16\textwidth,clip]{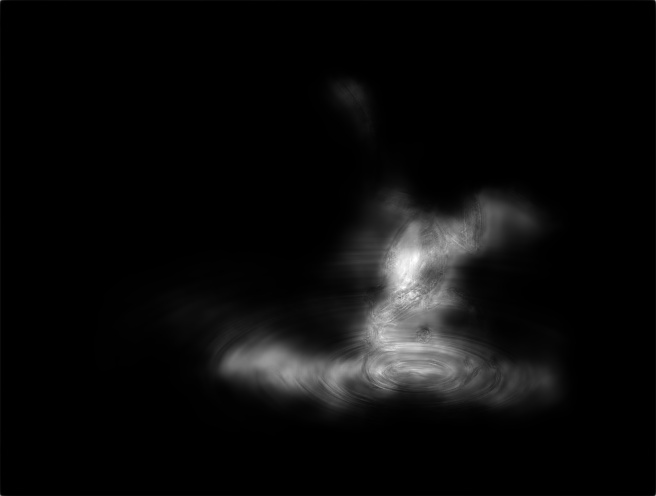}
        \includegraphics[width=0.15\textwidth, clip]{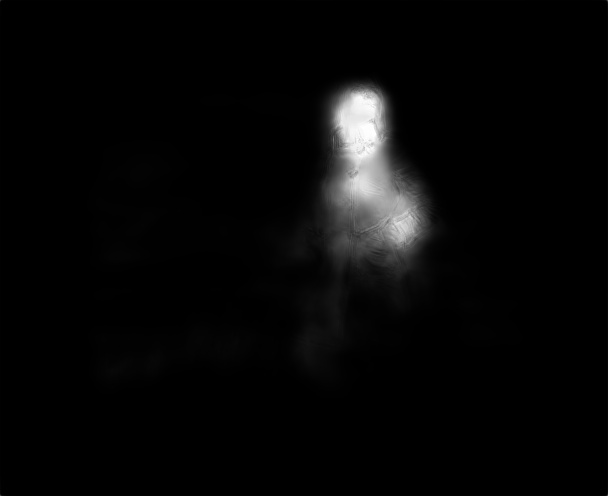} 
        \includegraphics[width=0.18\textwidth, clip]{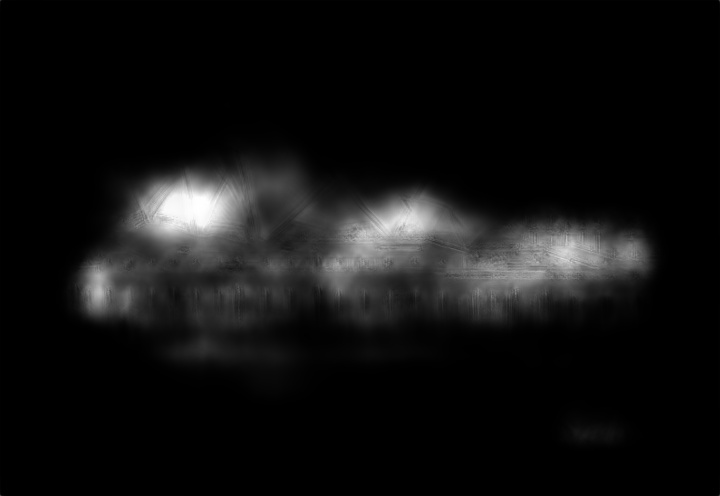}
        \includegraphics[width=0.15\textwidth, clip]{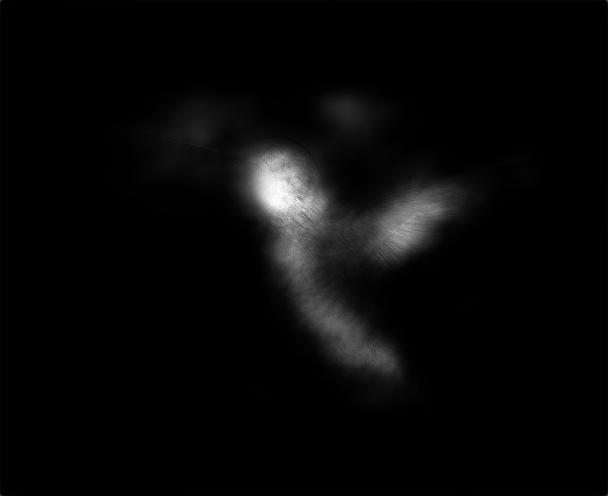} \label{fig:ava-saliency}} \\
    \subfigure[]{\includegraphics[width=0.16\textwidth,clip]{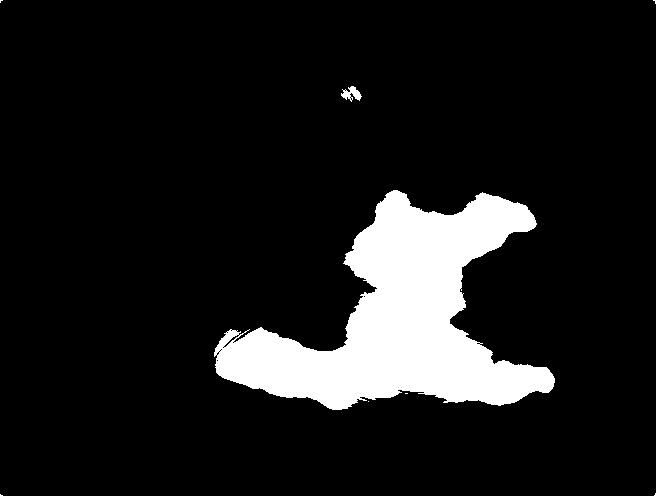}
        \includegraphics[width=0.15\textwidth, clip]{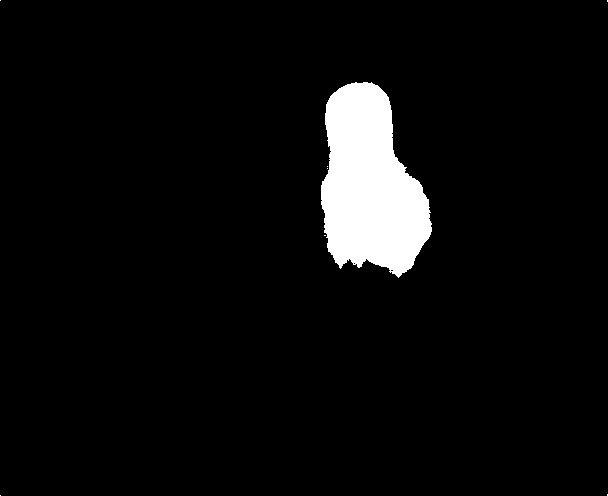} 
        \includegraphics[width=0.18\textwidth, clip]{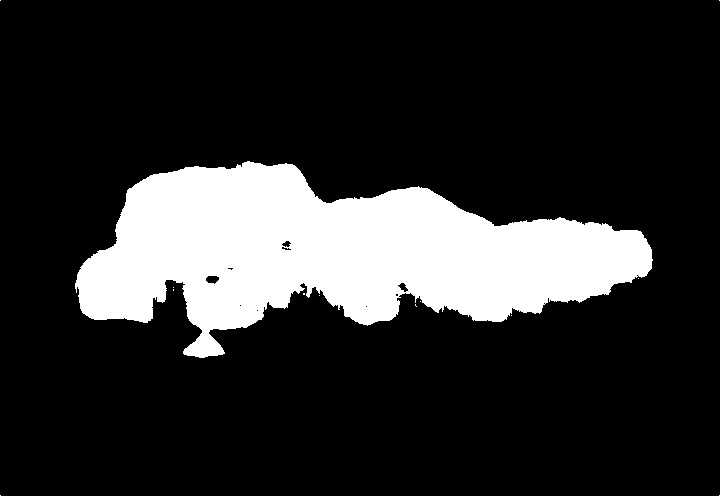} 
        \includegraphics[width=0.15\textwidth, clip]{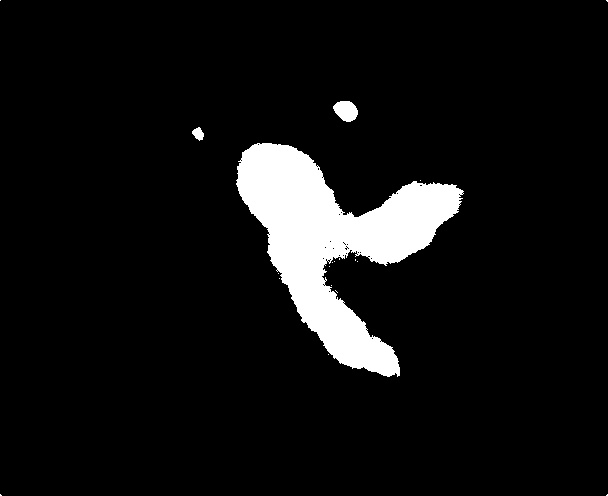} \label{fig:ava-binary}}
    \caption{Sample images along with their ground truths from AVA dataset. (a) Sample images  with high aesthetic scores from AVA dataset. (b) Corresponding saliency maps for sample images. (c) Binarized interested objects image for AVA samples where threshold is 0.12. \label{fig:ava-samples} } 
\end{figure*}

We conducted our experiments on the following four datasets.
\subsubsection{AVA database }
The AVA databae \cite{6247954}, which was originally designed for aesthetic visual analysis, gathered more than $250,000$ images from www.dpchallenge.com. Each image in AVA set contains plenty of meta-data, including multiple aesthetic scores from reviewers, semantic labels for over 60 categories, etc. In this work, we utilized AVA database to train the proposed end-to-end image cropping network, where only images whose average aesthetic scores were 
greater than or equal to 6 were selected for training, which resulted in a training set with $50,189$ high qualities images. Sample images from AVA dataset can be found in Figure \ref{fig:ava-origin}. \par

However, the AVA dataset was originally designed for aesthetic evaluation and only aesthetic scores were provided as the ground truths for each image. So in order to train the proposed neural networks with this dataset, the synthetic ground truths of salient objects image and offsets of final crop window with respect to the anchor region for each image were produced initially. \par

To detect the salient object for each image in AVA dataset, the method in \cite{7410395} was implemented, where the SALICON dataset \cite{7298710} was used to train a single branched DNN for saliency map detection. Based on the obtained saliency map, the binarized image can be calculated by using a simple thresholding approach with empirical threshold. In Figure \ref{fig:ava-saliency} and \ref{fig:ava-binary}, the corresponding salieny maps and binarized salient objects for the sample images from AVA dataset are shown. \par

For the ground truths of offsets between final cropping region and the anchor region, the method described in subsection \ref{sec:loss} was applied, where the anchor region for each training image was calculated on the synthetic saliency map.  \par

\subsubsection{FCD database } The FCD  dataset \cite{7926615} was constructed to facilitate the aesthetic cropping task, where thousands images were collected from Flickr and cleaned by annotators. For each cleaned iamge, the cropping area was labeled by professional photographers and validated by multiple professional annotators who had passed Human Intelligence Tasks qualification test. Only those images that were ranked as preferable by at least 4 professional annotators were selected in the final cropping dataset. In our experiments, $1,335$ images were used as development set and $334$ samples were applied for evaluation purpose among this database, . \par

\subsubsection{FLMS database } FLMS dataset \cite{Fang:2014:AIC:2647868.2654979} colleced $500$ images from Flickr and the best cropping areas of each image were manually annotated by 10 experienced editors. In this work, we used FLMS dataset to evaluate the cropping performance. \par

\subsubsection{CUHK-ICD database } Furthormore, to measure the proposed image cropping method, CUHK-ICD dataset \cite{6618974} was employed. In this dataset, $950$ images were captured by amateur photographers but cropped by 3 professional editors. All images in this dataset were used for evaluation in our work. \par


To quantitatively evaluate the cropping performance, the intersection over union (IoU) and boundary displacement error (BDE) were employed, where IoU is defined as:
	\begin{equation}
    	IoU = \frac{A' \cap \hat{A}}{A' \cup \hat{A}}
    \end{equation}
and BDE is defined by:
	\begin{equation}
    	BDE = \sum_{k=1}^4 \norm{B'_k - \hat{B}_k} \big/ 4.
    \end{equation}
Here, $A'$ means the ground truth of the cropping area, $\hat{A}$ represents the predicted cropping region, and $B'_k$ and $\hat{B}_k$ are the normalized boundary coordinates for ground truth and predicted crop windows, respectively.  \par

\subsection{Neural Networks Training}
Because the proposed image cropping system contained two main components conceptually, a corresponding three-stage training scheme was applied, where the U-shaped saliency map detection network $H(\mathcal{I}, \mathbf\Theta_s)$ and the regression network $G(\mathcal{I}, \mathbf\Theta_r)$ were trained sequentially and the entire network was fine-tuned afterwords. \par

To train the U-shaped network, the images from AVA database  were employed, where $50,189$ images with their synthetic saliency maps  were fed into the network for training. In this experiment, SGD optimization scheme was applied and the training rate was fixed to $1 \times 10^{-4}$ for 4 epochs. \par


Once the U-shape network was learned, the obtained weights were locked for the second stage training of regression network. To train this regression network, the same training images with high qualities from AVA dataset were fed into U-shaped network to create saliency maps, from which an initial anchor region can be estimated based on the proposed anchor region generation layer subsequently. Then, the coordinates of this anchor region were passed to RoI layer, where the corresponding features from U-shaped network were extracted and sent to the following fully connected layers, as illustrated in Figure \ref{fig:structure}. The pre-calculated ground truths for offsets were used to guide the training of regression network to predict the offset between anchor region and the final cropping rectangle. We used SGD optimizer with learning rate of $1 \times 10^{-4}$ for 6 epochs in this training stage. \par

Finally, we used training images from AVA dataset along with the synthetic binarized saliency maps and pre-calculated offsets ground truths to fine-tune the entire network from end to end. The SGD optimizer was used in this stage with learning rate of $1 \times 10^{-5}$ for 2 epochs and the U-shaped saliency map detection network and aesthetic area regression network have the same loss weight. \par

In this three-stage training phase, the input images were resized so that the shorter side of the image was 224 but the original aspect ratio was maintained.

\subsection{Parameters' Tuning}

    \begin{figure}
        \centering
        \includegraphics[width=0.7\columnwidth, clip]{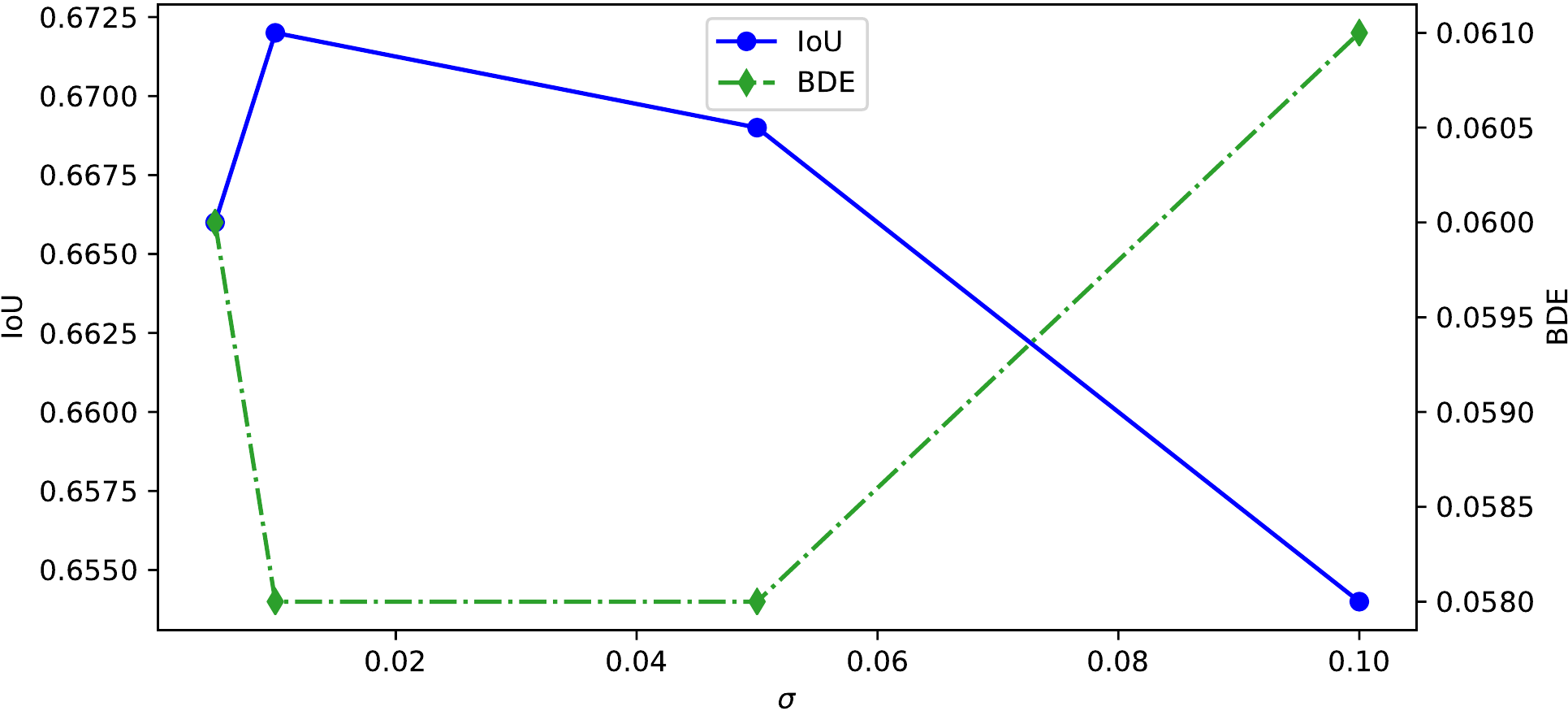}
        \caption{The cropping performance for different $\sigma$ on the development set. \label{fig:sigma}}
    \end{figure}

From subsection \ref{sec:enhancement}, we can see that the quality of interested objects within the obtained saliency map  can be enhanced by function $\rho(x; \sigma)$.  To investigate the impact from scale parameter $\sigma$ of function $\rho(x; \sigma)$ for the cropping system, we trained multiple end-to-end cropping networks using various values for $\sigma=\{0.005, 0.01, 0.05, 0.1\}$. We can notice that $\rho(x; \sigma)$ behaves like a soft binarizer which separates interested objects from the backgrounds within the saliency map. \par

To investigate the effect of $\sigma$ on final cropping results, the IoUs and BDEs were calculated on the development set from FCD database. In Fig. \ref{fig:sigma}, the IoU and BDE for different values of $\sigma$ on this development set are illustrated, where the parameter $\gamma$ for anchor region generation layer is set to be $3.0$. In this figure, it can be observed that the best cropping performance was achieved when $\sigma=0.01$, where IoU was 0.672 and BDE obtained 0.058. Therefore, in the final cropping system, we selected $\sigma=0.01$ for the function $\rho(x; \sigma)$ of soft binarization. \par

\subsection{Results Evaluation \& Analysis}
\subsubsection{Comparison with the state-of-the-art approaches}
To analyze the performance of the proposed end-to-end image cropping model, we compared the proposed method with other state-of-the-art cropping approaches, which were used as our baselines. \par
	\begin{table}[htp]
	    \scriptsize
    	\caption{IoUs and BDEs of different cropping approaches on CUHK-ICD dataset.\label{table:final_cuhk}}
        \centering          
        \begin{tabular}{ c | c c | c c | c c } 
          \hline
          \multicolumn{7}{c}{\bf{CUHK-ICD}} \\ \hline\hline
          \multirow{2}{*}{\bf{Approach}} & \multicolumn{2}{c|}{Photographer1} & \multicolumn{2}{c|}{Photographer2} & \multicolumn{2}{c}{Photographer3} \\
          {} & IoU & BDE & IoU & BDE & IoU & BDE \\ \hline          
            ATC \cite{Suh:2003:ATC:964696.964707} & 0.605 & 0.108 & 0.628 & 0.100 & 0.641 & 0.095 \\
            AIC \cite{7780430} & 0.469 & 0.142 & 0.494 & 0.131 & 0.512 & 0.123 \\
            LCC \cite{6618974} & 0.748 & 0.066 & 0.728 & 0.072 & 0.732 & 0.071 \\
            MPC \cite{6467466} & 0.603 & 0.106 & 0.582 & 0.112 & 0.608 & 0.110 \\
            ARC \cite{kong2016aesthetics} & 0.448 & 0.163 & 0.437 & 0.168 & 0.440 & 0.165 \\
            ABP-AA \cite{8365844} & 0.815 & 0.031 & 0.810 & 0.030 & 0.830 & 0.029 \\
            VFN \cite{Chen:2017:LCP:3123266.3123274} & 0.785 & 0.058 & 0.776 & 0.061 & 0.760 & 0.065 \\
            A2-RL \cite{8578953} & 0.802 & 0.052 & 0.796 & 0.054 & 0.790 & 0.054 \\
            Lu \emph{et al.} \cite{LU20191} & 0.827 & 0.032 & 0.816 & 0.035 & 0.805 & 0.036 \\
            Proposed & 0.822 & 0.031 & 0.815 & 0.034 & 0.802 & 0.035 \\ \hline
     \end{tabular}
    \end{table}

In table \ref{table:final_cuhk}, the IoUs and BDEs of the proposed cropping system on CUHK-ICD dataset are demonstrated, where other approaches are also listed and compared with. From these numbers, it can be found that the proposed method has competitive IoU and BDE performance on this evaluation set. The same metrics (IoU and BDE) were also utilized to evaluate the cropping systems on FLMS and FCD datasets, which are shown in table \ref{table:final_flms} and \ref{table:final_fcd}. From these two tables, we can see that the method proposed in this work provides better cropping performances than any other approaches on FLMS set and obtains the best BDE score on FCD dataset, which shows the effectiveness of the proposed cropping method. \par

	\begin{table}[htp]
	    \scriptsize
    	\caption{IoUs and BDEs of different cropping approaches on FLMS dataset.\label{table:final_flms}}
        \centering
        \begin{tabular}{ c | c  c}
          \hline
          \multicolumn{3}{c}{\bf{FLMS}} \\ \hline\hline
            \bf{Approach} & IoU & BDE \\ \hline
                VBC \cite{Fang:2014:AIC:2647868.2654979} & 0.740 & N/A \\
                ATC \cite{Suh:2003:ATC:964696.964707} & 0.720 & 0.063 \\
                AIC \cite{7780430} & 0.640 & 0.075 \\
                LCC \cite{6618974} & 0.630 & N/A \\
                MPC \cite{6467466} & 0.410 & N/A \\
                ABP-AA \cite{8365844} & 0.810 & 0.057 \\
                VEN \cite{wei2018good} & 0.837 & 0.041 \\
                A2-RL \cite{8578953} & 0.820 & N/A \\
                Lu \emph{et al.} \cite{LU20191} & 0.843 & 0.029 \\
                Proposed & 0.846 & 0.026 \\ \hline
          \end{tabular}
    \end{table}

	\begin{table}[htp]
	    \scriptsize
    	\caption{IoUs and BDEs of different cropping approaches on FCD dataset.\label{table:final_fcd}}
        \centering
        \begin{tabular}{ c | c  c}
          \hline
          \multicolumn{3}{c}{\bf{FCD}} \\ \hline\hline
            \bf{Approach} & IoU & BDE \\ \hline
                ATC \cite{Suh:2003:ATC:964696.964707} & 0.58 & 0.10 \\
                AIC \cite{7780430} & 0.47 & 0.13 \\
                RankSVM \cite{7926615} & 0.602 & 0.106 \\
                ARC \cite{kong2016aesthetics} & 0.484 & 0.140 \\
                MNA-CNN \cite{7780429} & 0.504 & 0.136  \\
                ABP-AA \cite{8365844} & 0.65 & 0.08 \\
                VEN \cite{wei2018good} & 0.735 & 0.072 \\
                VFN \cite{Chen:2017:LCP:3123266.3123274} & 0.684 & 0.084 \\
                Lu \emph{et al.} \cite{LU20191} & 0.659 & 0.062 \\
                Proposed & 0.673 & 0.058 \\ \hline
          \end{tabular}
    \end{table}

By analyzing the cropping results on three different datasets, we can find that the cropping performances are biased to the FCD dataset  mainly because the development set was constructed from it, and the ground truths from FCD dataset are not consistent with the annotated cropping areas in CUHK-ICD dataset, where the labeled cropping windows are relatively small compared to the salient objects in FCD dataset, whilst the ground truths from CUHK-ICD dataset tend to be large to cover entire image. \par

\begin{figure*}
        \centering
        \includegraphics[width=0.65\textwidth,clip]{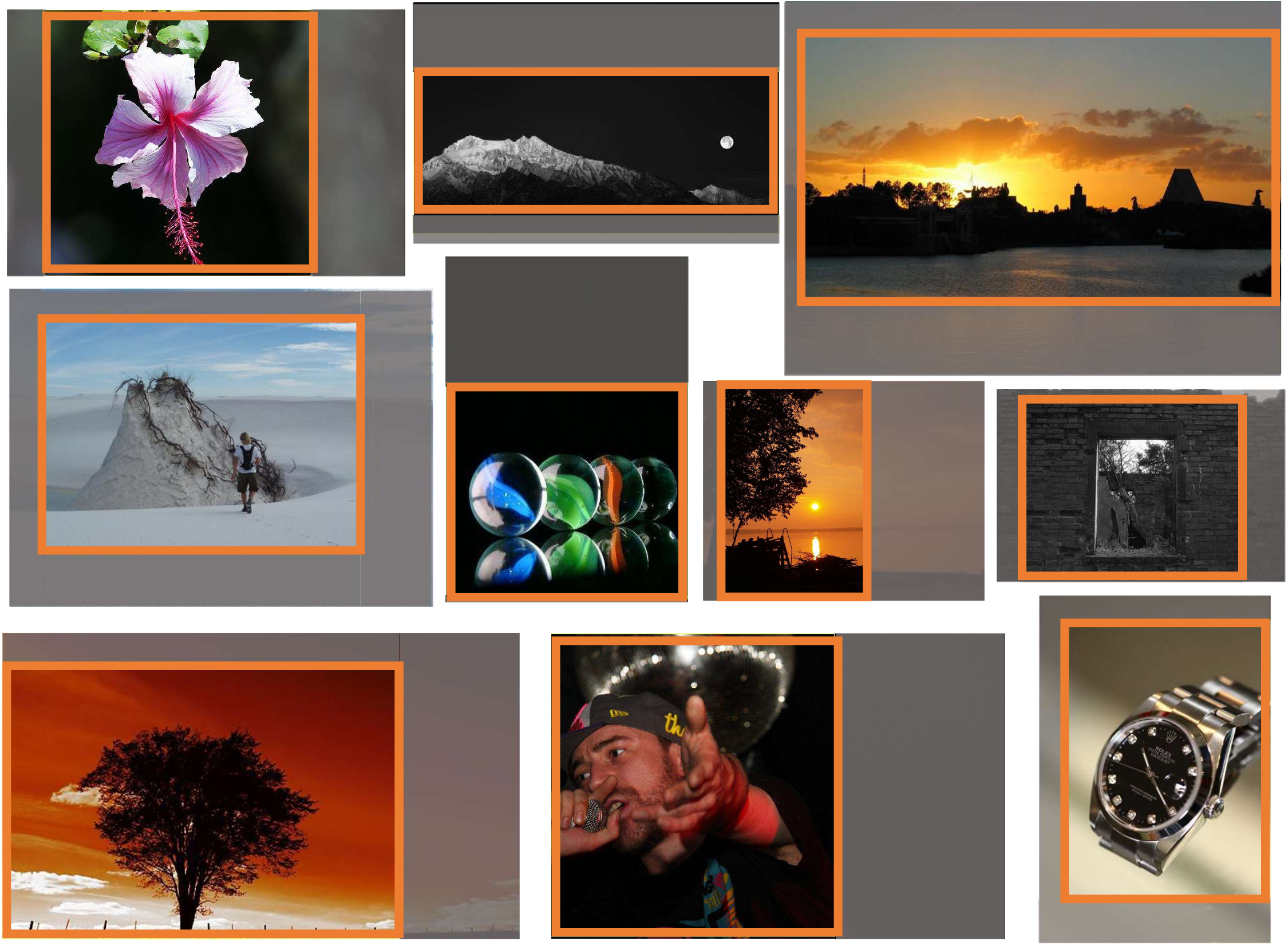}
        \caption{Cropping rectangle produced by the proposed system.}
        \label{fig:cropping_results}
    \end{figure*}

In Fig. \ref{fig:cropping_results}, multiple cropping results from the evaluation sets are demonstrated, where orange boxes represent the optimized cropping window predicted by the proposed system. From these images, we can see that the cropped images obtain better composition and aspect ratio than the original images, especially for those 
amateur captured low quality images, such as the center right scenery and bottom center portrait. \par

\subsubsection{Ablation test}

    \begin{table*}[h]
        \scriptsize
        \caption{The cropping performance using Salicon based saliency detection network on CUHK-ICD, FLMS and FCD datasets.\label{table:salicon}}
        \centering
        \begin{tabular}{c|c c|c c|c c|c c|c c}
            \hline
             \multirow{3}{*}{Method} & \multicolumn{6}{c|}{CUHK-ICD} & \multicolumn{2}{c|}{\multirow{2}{*}{FLMS}}  & \multicolumn{2}{c}{\multirow{2}{*}{FCD}}  \\ \cline{2-7}
                      & \multicolumn{2}{c|}{Photographer1} & \multicolumn{2}{c|}{Photographer2} & \multicolumn{2}{c|}{Photographer3} & & & \\ \cline{2-11} 
              & IoU & BDE & IoU & BDE & IoU & BDE & IoU & BDE & IoU & BDE \\ \hline\hline
            Salicon + Regression & 0.819 & 0.032 & 0.808 & 0.036 & 0.799 & 0.037 & 0.838 & 0.028  & 0.666 & 0.060 \\
            U-shaped + Regression & 0.825 & 0.032 & 0.820 & 0.034 & 0.806 & 0.036 & 0.845 & 0.028 & 0.664 & 0.060 \\ \hline
        \end{tabular} 
    \end{table*}

In the proposed image cropping framework, the U-shape based saliency map generation network can be easily replaced by other state-of-the-art saliency detection modules. Thus, in our experiments, we re-trained the SALICON saliency detection network, which was introduced in \cite{7410395} and applied to generate the synthetic ground truth for AVA dataset in section \ref{sec:dataset}, to detect the saliency maps for test images and consequently feed them into the anchor region generation layer and aesthetic area regression network to produce the final cropping window. In table \ref{table:salicon}, the overall cropping performances by combining the SALICON saliency map detection network and the proposed aesthetic area regression network are listed, where we can see it provides similar cropping results compared with the U-shape based saliency detection network, which shows the generalization capability of the proposed framework. \par

Need to note that in the ablation test, in order to avoid the size of feature maps extracted by SALICON saliency map detection network being too small, the input images of the neural networks were resized to ensure the shorter side of the image was 512 with the original aspect ratio. \par

By analyzing three tables and the structure of SALICON network and U-shaped network, it can be concluded that the cropping performance differences between these two saliency detection modules rely on the resolution of extracted features from these two networks. For the SALICON network, it applies the VGG-16 to extract down-sampled feature maps for images, which provides coarse details of the salient objects. But U-shaped saliency detection network extracts the feature map whose size is the same as the input image, that maintains more details of the interested object. Therefore, for the cleaned high resolution images, such as photos from AVA dataset, U-shaped saliency detection network tends to extract more pleasant features of the interested objects in the image to help the cropping task. And for the noisy low quality images, SALICON network acts more like a noise supresser to extract smoothed features to boost the cropping performance, as we observed from FCD dataset. \par

\subsubsection{Investigation of image's size and aspect ratio}

    \begin{table*}[h]
        \scriptsize
        \caption{The cropping performance for different aspect ratio and image size on CUHK-ICD, FLMS and FCD datasets.\label{table:ratio}}
        \centering
        \begin{tabular}{c|c c|c c|c c|c c|c c}
            \hline
             \multirow{3}{*}{Input size} & \multicolumn{6}{c|}{CUHK-ICD} & \multicolumn{2}{c|}{\multirow{2}{*}{FLMS}}  & \multicolumn{2}{c}{\multirow{2}{*}{FCD}}  \\ \cline{2-7}
                      & \multicolumn{2}{c|}{Photographer1} & \multicolumn{2}{c|}{Photographer2} & \multicolumn{2}{c|}{Photographer3} & & & \\ \cline{2-11} 
              & IoU & BDE & IoU & BDE & IoU & BDE & IoU & BDE & IoU & BDE \\ \hline\hline

            $224 \times 224$ & 0.825 & 0.031 & 0.818 & 0.034 & 0.805 & 0.036 & 0.840 & 0.028  & 0.672 & 0.059 \\
            $384 \times 384$ & 0.827 & 0.031 & 0.817 & 0.034 & 0.804 & 0.036 & 0.843 & 0.028 & 0.670 & 0.059 \\
            $512 \times 512$ & 0.828 &  0.031 & 0.822 &  0.034 &  0.806 &  0.036 & 0.842 & 0.028 & 0.665 & 0.061 \\
            $\min(w, h)=224$ & 0.822 & 0.031 & 0.815 & 0.034 & 0.802 & 0.035 &  0.846 &  0.026 &  0.673 &  0.058 \\
            $\min(w,h)=384$ & 0.823 & 0.032 & 0.818 & 0.034 & 0.804 & 0.036 & 0.844 & 0.027 & 0.670 & 0.059 \\
            $\min(w,h)=512$ & 0.825 & 0.032 & 0.820 & 0.034 & 0.806 & 0.036 & 0.845 & 0.028 & 0.664 & 0.060 \\ \hline
        \end{tabular} 
    \end{table*}

In many other research articles, it is claimed that the aesthetic quality of images is highly relied on the size or aspect ratio of the images \cite{8099567, 7780429}. Thus, we carried out several experiments to investigate cropping performance with different image size and aspect ratio. In these experiments, we trained three models by keeping the original aspect ratio of the images but resizing the image till the shorter side of the image is 224, 384 or 512. Other three models were trained by resizing the image to square, whose size is $224 \times 224$, $384 \times 384$ or $512 \times 512$, respectively. \par

In table \ref{table:ratio}, we list the IoUs and BDEs of different models with various input image size on the three public test sets. From this table, it can be seen that both IoUs and BDEs for different input size of images have similar performances and no significant difference can be found between these models, which means the proposed image cropping model is insensitive to the size and aspect ratio of the input image. \par

By digging into table \ref{table:ratio}, we observed that the overall IoU and BDE scores on CUHK-ICD dataset were getting better with larger input image size of neural networks, whilst the cropping performance was degraded on FCD dataset with larger input image size. The main reason of this phenomenon is that the images in the FCD dataset were collected from Flickr's website, containing more irrelevant background noises than the training dataset and other two evaluation datasets. With a larger input size, more detailed features of images from FCD dataset, including non-interested background noises, can be discovered by the neural networks. But these features of noises cannot be effectively represented by the neural network which was trained based on the clean images from AVA dataset, and can be easily mis-represented as objects' features. This causes the proposed cropping method tending to generate larger crop windows to include more details when the input image size is big, which degrades the performance of the system on FCD dataset. But for the FLMS dataset, each test image had multiple annotated ground truths and the best cropping result was calculated using the ground truth which provided best performance. So, FLMS set is less sensitive to the neural network's input image size. With regard to CUHK-ICD dataset, it was constructed by high aesthetic quality images similar to AVA dataset, which in turn can be sufficiently embedded by the proposed cropping networks with larger input image size to attain better cropping performance. \par

Compromised by the cropping performance across three evaluation sets and the computation efficiency, it is preferable to resize the input image such that the shorter side is 224 for the proposed cropping approach. \par

\subsubsection{Efficiency analysis}
As one of the main contributions of this work is to use an end-to-end neural network to accomplish the image cropping task, without iteratively evaluating multiple candidates' aesthetic qualities, which has lower computational cost. So we measured time efficiency of the proposed system with different input size on the FLMS set, where the experiments were implemented with Keras on a server with Intel(R) Xeon(R) E5-2620 CPU @ 2.10GHz, 64Gb Memory and Nvidia 2080Ti GPU. In table \ref{table:speed}, the average processing time for each test image with different input size is shown. From this table, we notice that when the input image  is resized to $224 \times 224$, the overall time for image cropping is less than 20ms. Thus, the proposed system can reach 50fps on average for real-time processing, which shows its high efficiency. \par

Furthermore, by comparing the time efficiency with the cropping method presented in \cite{LU20191} which is relied on a brute force search algorithm \cite{7780430}, we can find that based on the introduced anchor region generation layer, the proposed image cropping system can effectively produce a seed region that contains the saliency area and re-use the features from the saliency map generation network, such that it is five times faster than the method in \cite{LU20191}.

    \begin{table}[htp]
        \scriptsize
        \caption{The time efficiency test of the proposed cropping system on FLMS dataset. \label{table:speed}}

        \centering
        \begin{tabular}{c|c }
            \hline
            Input size &  \makecell{Crop window \\ prediction time (ms)} \\ \hline
            $224 \times 224$ &  19 \\
            $384 \times 384$ &  35 \\
            $512 \times 512$ &  55  \\
            $\min(w, h)=224$ &  25 \\
            $\min(w,h)=384$ &  46 \\
            $\min(w,h)=512$ &  71 \\ \hline
        \end{tabular} 
    \end{table} 

\subsection{Subjective analysis}
Because image's aesthetics is difficult to represent from the subjective perspective, such that different person might have different views for the same cropping results based on their tastes, education backgrounds, etc. So, in our work, a subjective comparative experiment was carried out. \par

\begin{figure}
    \centering
    \includegraphics[width=0.7\columnwidth,clip]{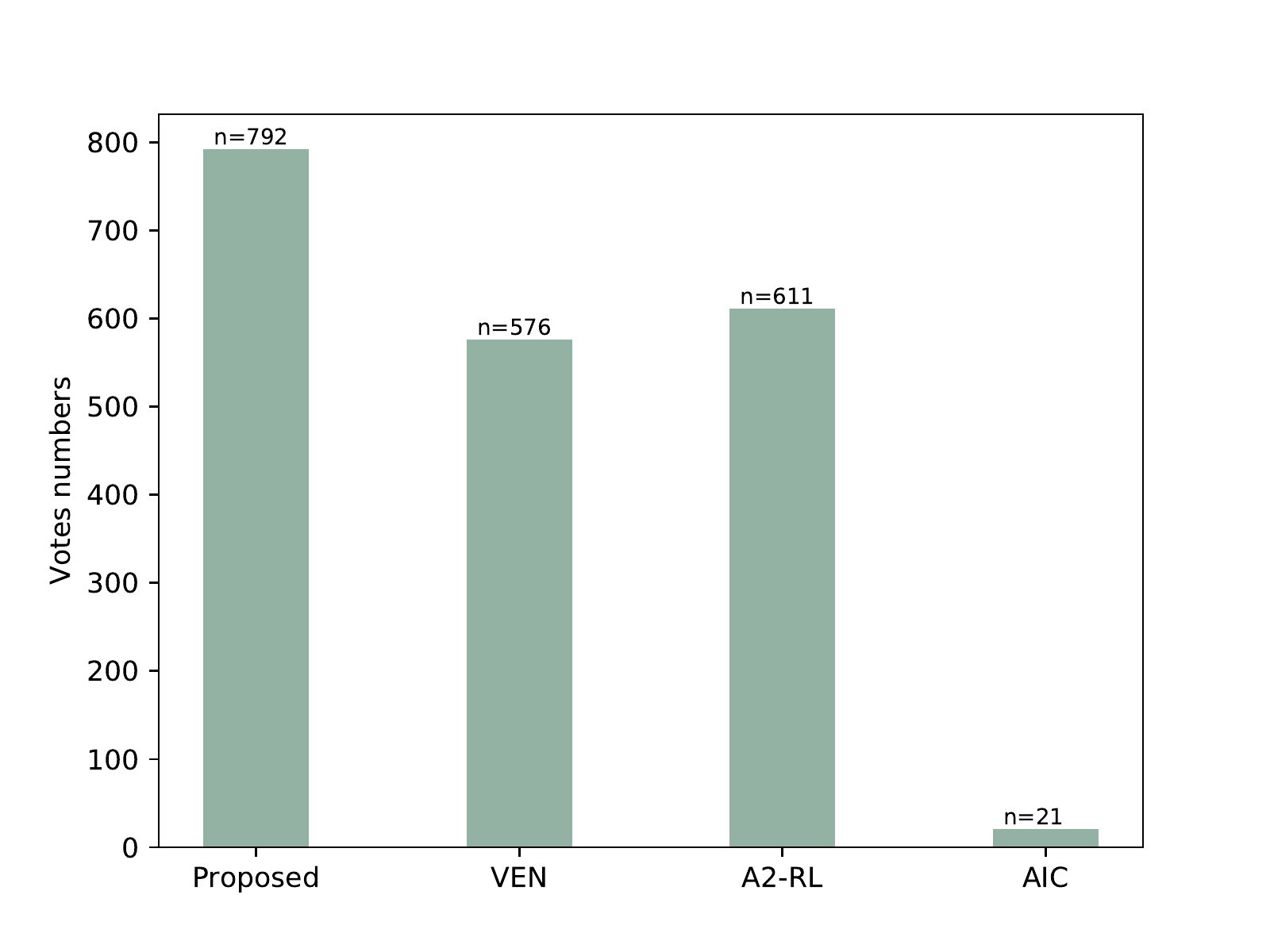}
    \caption{Votes recieved from users for different state-of-the-art cropping methods.}
    \label{fig:subjective}
\end{figure}

In this experiment, 200 images were randomly collected from three (CUHK-ICD, FLMS and FCD) test sets. For each image, the proposed cropping method, along with the algorithms AIC \cite{7780430}, A2-RL \cite{8578953} and VEN \cite{wei2018good} was employed to obtain four cropping results. Then, 10 users were recruited, including 5 males and 5 females. All users had no prior knowledge of the experiment content and the datasets. For each participant, the four cropping results of each test image were presented, where the order of the cropping images was randomized and the users were asked to vote the most pleasing one in terms of their aesthetics. Finally, 2000 votes from 10 participants were received and shown in figure \ref{fig:subjective}.

As can be seen from this figure, the proposed method had gained most votes ($792/2000$) among four state-of-the-art cropping approaches, which shows the proposed system provided more pleasing cropping results than the other methods in respect to the aesthetics.\par
    
\subsection{Failure Case Study}
Although the proposed image cropping approach works well on the majority of testing images, several failure cases can be found in the evaluation, which can be categorized into two types of errors broadly. \par

\begin{figure}
        \centering
        \subfigure[Failure examples from FLMS dataset.]{\includegraphics[width=0.8\columnwidth,clip]{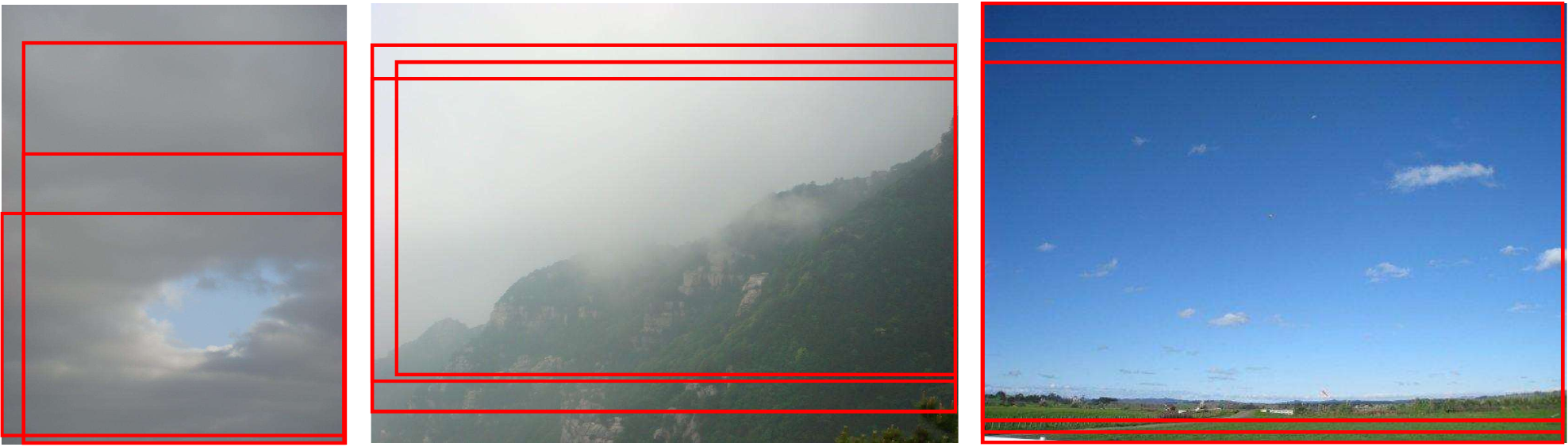} \label{fig:flms_failure}}
        \subfigure[Failure cases from FCD dataset.]{\includegraphics[width=0.8\columnwidth,clip]{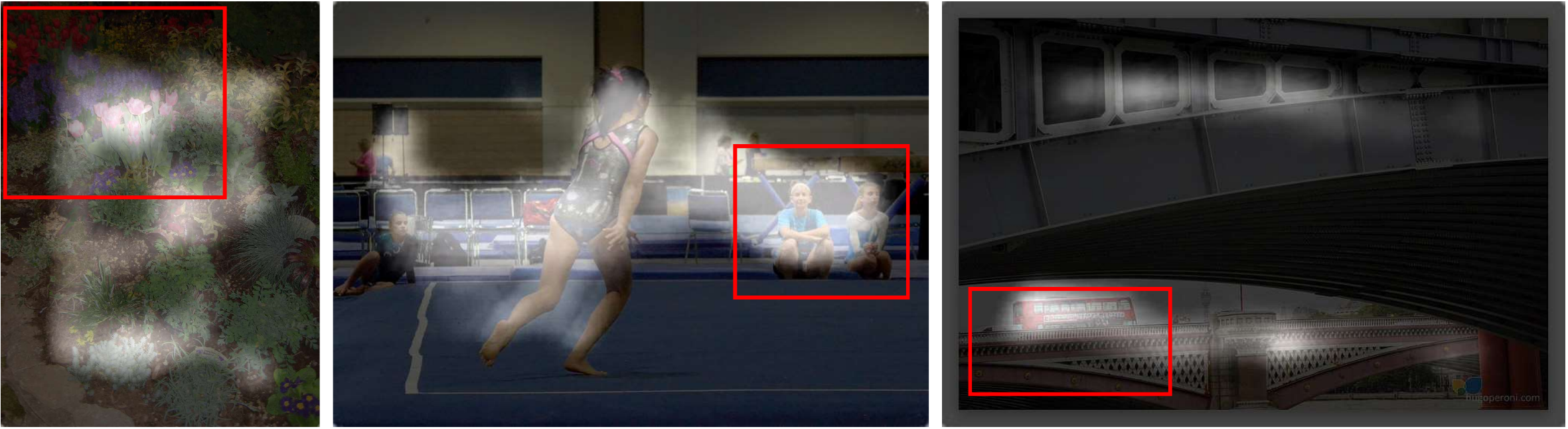} \label{fig:fcd_failure}}
        \caption{Failure examples from the evaluation sets. (a) Failure samples from FLMS dataset, where red boxes are cropping windows by annotators. (b) Failure images from FCD dataset, where red boxes are ground truth and light areas are the detected saliency maps.}
        \label{fig:failure_cases}
    \end{figure}

The first type of failures are mostly from the FLMS dataset, as shown in the Fig. \ref{fig:flms_failure}, where only spurious texture regions exist in the image and it is hard to find enough salient pixels to determine the interested objects in the image. In our implementation, if no visual fixation is found, we use center areas that cover the $70\%$ of entire image as our anchor region to feed into regression network to obtain the final cropping rectangle. The other type of failure case can be seen in FCD dataset, where multiple interested objects are detected by the saliency map detection network, as shown in Fig. \ref{fig:fcd_failure},  but only partial of these saliency area is included into the ground truth and most parts are missing, which causes the low IoUs and BDEs. 

\section{Conclusion} \label{sec:conclusion}
In this paper, an end-to-end automatic image cropping system is proposed to learn the relationship between the objects of interests and the areas with high aesthetic scores in an image through a DNN. Conceptually, the saliency map is initially detected by using a U-shaped neural network, which is then passed into a soft binarization layer to separate objects from the background. Based on this enhanced saliency map, an anchor region is determined by the proposed anchor region generation layer, which is fed into a ROI pooling layer and following dense layers along with the features of the interested objects, to predict the optimal cropping region with high aesthetic scores. \par

The proposed algorithm outperforms other state-of-the-art cropping method with regard to IoU and BDE metrics. Moreover, because the proposed approach finds the final cropping areas based on the hidden relationship between interested objects and areas with high aesthetics quality through neural networks, which avoids to iteratively evaluate  multiple cropping candidates, high processing efficiency is achieved compared with other approaches. \par

Our future research will be exploring other cropping metric instead of IoU and BDE to measure the performance of aesthetics based cropping system.

\ifCLASSOPTIONcaptionsoff
  \newpage
\fi



\bibliographystyle{IEEEtran}
\bibliography{refs}

\end{document}